# Research on Cervical Cancer p16/Ki-67 Immunohistochemical Dual-Staining Image Recognition Algorithm Based on YOLO

Xiao-Jun Wu, Cai-Jun Zhao, Chun Meng, Hang Wang

**Abstract**: The p16/Ki-67 dual staining method is a new approach for cervical cancer screening with high sensitivity and specificity. However, there are issues of mis-detection and inaccurate recognition when the YOLOv5s algorithm is directly applied to dual-stained cell images. This paper Proposes a novel cervical cancer dual-stained image recognition (DSIR-YOLO) model based on an YOLOv5. By fusing the Swin-Transformer module, GAM attention mechanism, multi-scale feature fusion, and EIoU loss function, the detection performance is significantly improved, with mAP@0.5 and mAP@0.5:0.95 reaching 92.6% and 70.5%, respectively. Compared with YOLOv5s in five-fold cross-validation, the accuracy, recall, mAP@0.5, and mAP@0.5:0.95 of the improved algorithm are increased by 2.3%, 4.1%, 4.3%, and 8.0%, respectively, with smaller variances and higher stability. Compared with other detection algorithms, DSIR-YOLO in this paper sacrifices some performance requirements to improve the network recognition effect. In addition, the influence of dataset quality on the detection results is studied. By controlling the sealing property of pixels, scale difference, unlabelled cells, and diagonal annotation, the model detection accuracy, recall, mAP@0.5, and mAP@0.5:0.95 are improved by 13.3%, 15.3%, 18.3%, and 30.5%, respectively.

## 0 Introduction

Cervical cancer stands as one of the most prevalent malignant tumors affecting women, posing a significant threat to their health and quality of life. In 2020, the World Health Organization (WHO) initiated the Global Cervical Cancer Elimination Initiative[1], aimed at accelerating efforts towards its eradication. The Chinese government, demonstrating a strong commitment to women's health, outlined in the "Healthy China 2030" blueprint issued in 2016, the objective of enhancing screening rates and early diagnosis for common gynecological diseases. Further, the "Program for the Development of Chinese Women (2021-2030)" underscores the ambition to augment capabilities in cervical cancer prevention, aspiring to elevate screening coverage to over 70%, thereby safeguarding the health of eligible females[2]. However, only one in five eligible Chinese women have undergone cervical cancer screening within a three-year period [3], a figure that falls drastically short of the 70% target. A fundamental impediment to the expansion of cervical cancer screening programs is the scarcity of pathologists, compounded by issues related to female participation and accessibility, which curtail the reach of such initiatives in China. Training a sufficient number of pathologists necessitates substantial investments of time and finances. Consequently, devising novel strategies to bolster cervical cancer screening capabilities, in alignment with current needs and the WHO's objectives, is imperative.

In recent years, the p16/Ki-67 dual-staining technique has garnered significant attention as a novel approach to cervical cancer screening. By concurrently staining p16 and Ki-67 proteins in cervical cells, this method enhances the accuracy in assessing cellular proliferation and abnormal changes, thereby improving the early detection rate of cervical cancer. Typically, p16 is expressed in both the cytoplasm and nucleus, stained in magenta, while Ki-67 is nuclear, marked in brownish-yellow. Under the same microscopic focus, when both cytoplasmic (magenta) and nuclear (brownish-yellow) staining coexist in a single cell, it is deemed a dual-positive cell. In liquid-based thin layer preparations, the identification of one or more dual-positive cells or clusters thereof warrants a report of p16/Ki-67 dual positivity.

Artificial Intelligence (AI)-assisted screening emerges as a practical alternative to overcome the shortage of pathologists and enhance the efficiency of cervical cancer screening in China. The integration of AI-assisted detection with p16/Ki-67 dual-staining has the potential to augment the precision and efficiency of cervical screening without necessitating substantial additional investments in medical personnel training and accreditation, nor causing delays due to such training. This could prove particularly beneficial for remote and rural regions of China where qualified pathologists are scarce. As technology advances and clinical data accumulates on HPV and cervical cancer diagnoses, AI-assisted diagnostic systems are poised to become increasingly accurate, efficient, and cost-effective for population-wide screening. AI-augmented cytological testing for cervical cancer can dramatically elevate screening rates in China, propelling progress towards the WHO's ambitious goal of cervical cancer elimination.

# 1 YOLOv5 Structure

The YOLOv5 algorithm, while demonstrating remarkable performance on general datasets such as COCO and VOC, does not fully cater to the specific requirements of the dual-stained cervical cell image dataset constructed in this study. Building upon the YOLOv5 framework, a series of network enhancement strategies tailored to the characteristics of cervical dual-stained cell images have been proposed.

An improved algorithmic network structure, Combining the Swin-Transformer module, GAM attention mechanism, multi-scale feature fusion and EIoU loss function, efficient detection of cervical cancer double-stained cells is achieved, depicted in Figure 1, has been devised to better suit the task at hand.

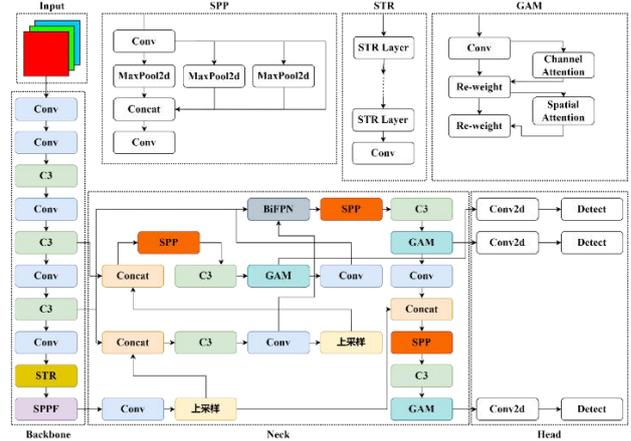

Figure 1: The structure of DSIR-YOLO

## 1.1 Integration of Swin-Transformer Module

The Swin-Transformer module, whose architecture is shown in Figure 2, constitutes a pivotal enhancement. It primarily comprises Layer Normalization (LN), Window-based Multi-head Self-Attention (W-MSA), Shifted Window-based Multi-head Self-Attention (SW-MSA), and Multi-Layer Perceptrons (MLPs). These components collectively contribute to enhancing the model's stability, expressive power, and efficiency. Specifically, LN aids in preventing gradient vanishing or explosion, MLPs introduce non-linearity, and residual connections preserve original input information, thereby enriching the model's learning capacity.

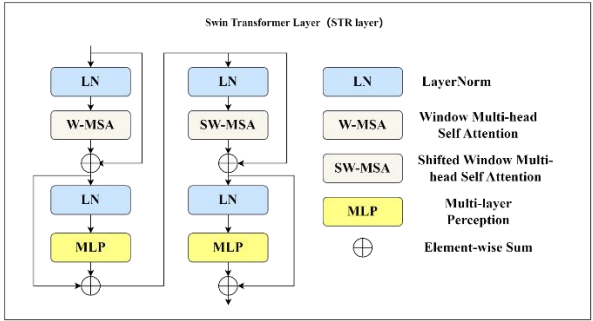

Figure 2: Structure of the Swin-Transformer Module

The Swin-Transformer has proven its efficacy across various visual tasks, particularly excelling in capturing both global and local information within images. For instance, Dee W et al[4] employed a Swin-Transformer-based computer vision architecture to directly predict the mechanisms of action for ten kinase inhibitor compounds from raw cell response images, significantly improving prediction accuracy. In this work, the Swin-Transformer module is integrated to replace the Backbone component of YOLOv5, with the aim of enhancing the model's adaptability to the diversity present in cellular images, thereby boosting its performance in the context of cervical cancer screening using dual-stained cell imagery.

## 1.2 Global Attention Mechanism

The Global Attention Mechanism (GAM) [5] is a holistic attention mechanism designed to retain and enhance interactions between the channel and spatial dimensions. As shown in Figure 3, GAM primarily employs Spatial Attention Module (SAM) and Channel Attention Module (CAM) to boost the feature extraction capability of YOLOv5. SAM focuses on capturing spatial dependencies within feature maps, whereas CAM emphasizes inter-channel relationships, collectively elevating the expressiveness and discriminative power of these maps.

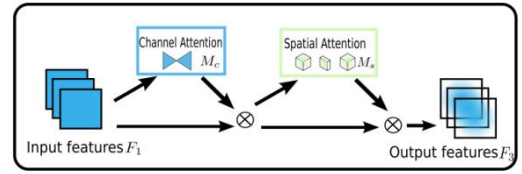

Figure 3: Framework of the Global Attention Mechanism

## 1.3 Multi-Scale Feature Fusion

Multi-scale feature fusion, a strategy that integrates features from different scales [6], augments the model's ability to discern cell sizes and shapes. By effectively capturing context and dependencies across various image scales, this approach enables accurate detection of objects of differing sizes [7].

In this adaptation, the Concatenation (Concat) module after downsampling is replaced with a module akin to the Bi-directional Feature Pyramid Network (BiFPN). BiFPN facilitates bidirectional feature fusion, combining top-down and bottom-up pathways, and employs weighted residual connections along with fast normalization to enhance fusion efficiency and stability.

To further preserve and extract multi-scale cellular features, a Spatial Pyramid Pooling (SPP) structure is incorporated post neck down-sampling, as depicted in Figure 4[8]. In YOLOv5s, concatenated features from different levels or scales, concatenated along the channel dimension via Concat, increase the dimensionality and diversity of features. These features then pass through the C3 structure, designed to deepen the network and expand the receptive field, thereby improving feature extraction. The integration of the SPP module between Concat and C3 further expands the receptive field of the concatenated feature maps, ensuring adequate feature extraction for cells of all sizes present in the dataset and adapting well to the variable shapes of cells. Additionally, by conducting multi-scale pooling operations on the feature maps before they are halved in size, SPP prior to C3 enhances spatial information capture, boosting the regression accuracy for cell recognition.

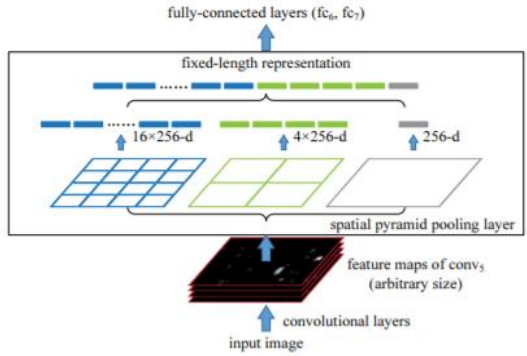

Figure 4: Diagrammatic Representation of the Spatial Pyramid Pooling Structure

## 1.4 EIoU Loss Function

In place of the conventional CIoU (Complete Intersection over Union) loss function for bounding box regression, the Focal-EIoU (Efficient Intersection over Union) loss function is adopted. EIoU[9] is a loss function grounded in the concept of Intersection over Union (IoU), which comprehensively assesses three geometric aspects of bounding box regression: overlapping area, central point alignment, and aspect ratio, encapsulated in Equation 1. By incorporating these factors, the EIoU loss effectively enhances both the precision and robustness of boundary box localization.

$$L_{EIoU} = L_{IoU} + L_{dis} + L_{asp}$$
$$= 1 - IoU + \frac{\rho^2(b, b^{gt})}{c^2} + \frac{\rho^2(w, w^{gt})}{C_w^2} + \frac{\rho^2(h, h^{gt})}{C_h^2} \quad (1)$$

# 2 Dataset Construction and Optimization

## 2.1 Experimental Data

Eleven electronic pathological images of p16/Ki-67 dual-stained liquid-based cervical cell samples were utilized, provided by Fuzhou Maxin Biotechnology Development Co., Ltd. These images encapsulate diverse cell morphologies, densities, and distributions, thereby ensuring representativeness and variety. The images were segmented into 1024×1024 TIFF format image patches, yielding a total of 7,662 patches. Subsequently, the dataset was randomly divided into training and validation sets at an 8:2 ratio for network training and evaluation purposes, respectively.

## 2.2 Data Processing

State-of-the-art biomedical image analysis algorithms necessitate high-quality dataset annotations, as highlighted in recent research[10, 11]. Given the scarcity of medical image data and potential variations in interpretation among different evaluators, annotating these datasets introduces an element of uncertainty. To ensure accuracy and reliability, annotation tasks are conducted under the guidance of medical experts. However, the scarcity of such experts and the potential influence of individual styles in their annotations may compromise dataset consistency[12]. Conversely, while involving non-experts, like computer scientists, could enhance annotation consistency due to standardized procedures, their lack of domain knowledge, such as in pathology[13], might lead to incomplete or inaccurate datasets. Hence, the annotation of medical image datasets requires interdisciplinary expertise to optimize dataset quality. Enhancing the annotation workflow is pivotal not only for boosting model performance but also for securing its reliability and effectiveness.

Deep learning models necessitate substantial training data, yet no extensive dataset is publicly available for dual-staining in cervical cells. Consequently, data augmentation techniques, including Mosaic augmentation, MixUp augmentation, and random horizontal flipping, have been employed to expand the training dataset. This strategy bolsters the model's generalization capacity and robustness.

### 2.2.1 Enhancing Pixel Sealability

Refinement of bounding boxes is achieved by adjusting their positions and shapes to better conform to the contours of target objects,

thereby minimizing irrelevant or interfering information within the boxes. Inspection of the initial dataset revealed some cells with imprecise annotations, characterized by noticeable gaps between the bounding boxes and the cells themselves. A comparison of cell annotations before and after dataset optimization, as depicted in Figure 5, illustrates how the optimized bounding boxes are more compact and closely fit the target cell profiles, enhancing the network's learning efficiency.

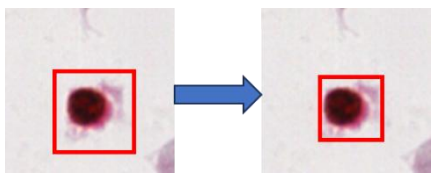

Figure 5: Illustration of Enhanced Pixel Sealability

### 2.2.2 Mitigating Annotation Scale Discrepancies

As shown in Figure 6, prior to optimization, annotations of cell clusters were often oversized compared to individual cells, creating a significant scale disparity between single-cell and cluster annotations. This discrepancy impedes the model's ability to strike a balance across scales during training, hindering the extraction of effective feature information. To address this, during optimization, larger cell clusters are subdivided into multiple smaller bounding boxes, reducing extraneous information and interference.

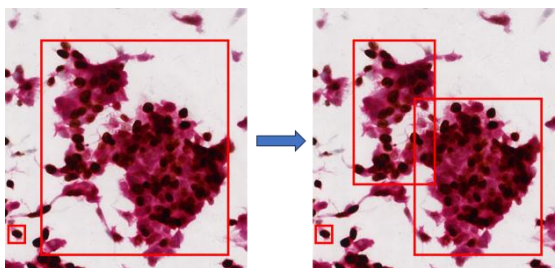

Figure 6: Reduction of Annotation Box Scale Discrepancy

### 2.2.3 Reducing Unlabeled Cells

Isolated positive cells, as seen in Figure 7, frequently surround large cell clusters without direct association. These cells are surrounded by extensive background and negative cells, which can confuse the network's recognition, leading to missed detections. By individually annotating these peripheral positive cells with distinct bounding boxes, issues of small cell omission due to overlay avoidance are mitigated. This approach heightens the network's sensitivity and precision for small cells, thereby improving detection coverage and completeness.

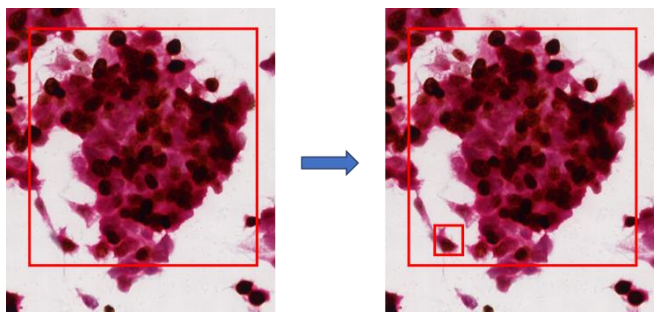

Figure 7: Reduction of Unlabeled Cells

### 2.2.4 Improving Annotations of Large-Angle Cell Clusters

For objects with substantial angular orientations, such as elongated cell clusters (depicted in Figure 8), their bounding boxes often encompass a much larger area than the actual object, incorporating excessive background and unrelated cells. Such bounding boxes undermine feature learning for positive cells by assigning equal importance to all pixels within, disregarding the shape and position of the cell cluster. Adopting a similar strategy to resolving multi-scale issues, large clusters are broken down into smaller ones, ensuring each bounding box is appropriately sized, minimizing superfluous information, and freeing positive cells outside the cluster boundary, thus increasing the ratio of informative content within the boxes.

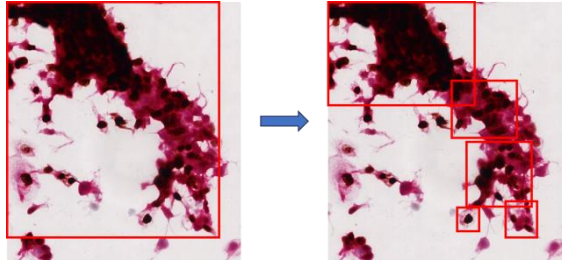

Figure 8: Avoidance of Whole Annotation for Large-Angle Cell Clusters

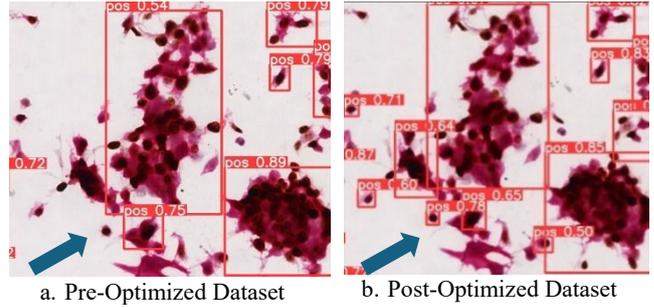

a. Pre-Optimized Dataset    b. Post-Optimized Dataset

Figure 9: Improved Dataset Quality Elevates Network Recognition

## 2.3 Impact of Dataset Quality on Recognition Performance

The unmodified YOLOv5s model was employed to detect dual-stained cell datasets both before and after optimization, with precision (P), recall (R), and mean Average Precision (mAP) across IoU thresholds from 0.5 to 0.95 as evaluation metrics. As shown in Table 1, enhancing the annotation quality of the dataset significantly improved the model's recognition performance.

Table 1: Model Performance Comparison Under Different Dataset Qualities

| Dataset Quality | P/% | R/% | $mAP_{@0.5:0.95}$/% |
|---|---|---|---|
| Before Optimization | 72.4 | 69.0 | 38.3 |
| After Optimization | 85.7 | 84.3 | 68.8 |

Figure 9 vividly demonstrates the impact of dataset annotation quality on model performance. As indicated by the arrows, post-optimization notably reduces the miss-detection of positive cells around cell clusters and enhances the accuracy in identifying small cells. These findings validate the crucial role of improved dataset quality in enhancing the efficacy of YOLOv5 in recognizing dual-stained positive cells in cervical cancer.

## 3 Model Recognition Improvement and Analysis

### 3.1 Experiment Environment

The deep learning experiments were conducted on a Windows 11 system equipped with an Intel(R) Core(TM) i7-12700F CPU, NVIDIA RTX 3070 (8GB) GPU, and 32GB RAM. The PyTorch 1.12 framework with Python 3.7 was used, accelerated by CUDA 12.2.

The primary hyperparameters for the deep learning experiments included: a training epoch count of 100, an initial learning rate of 0.01 reducing to a minimum of 0.0001. Stochastic Gradient Descent (SGD) served as the optimizer, with weight decay set at 0.0005. Testing was performed using non-maximum suppression threshold 0.60, confidence threshold 0.001, and prediction probability threshold 0.5.

### 3.2 Evaluation Metrics

Model performance metrics, specifically precision, recall, and mean Average Precision (mAP) at IoU thresholds of 0.5 and from 0.5 to 0.95, were employed to assess the performance of cervical dual-stain cell detection. Calculation methods are derived from the confusion matrix of binary classification, as shown in Figure 10. In YOLOv5, values in the confusion matrix are determined based on the Intersection over Union (IoU) calculation between detected and annotated bounding boxes.

|  | Prediction | |
|---|---|---|
| Confusion Matrix | Positive | Negative |
| True Positive | True Positive | False Negative |
| True Negative | False Positive | True Negative |

Figure 10: Confusion Matrix

$$\text{Precision} = \frac{T_P}{T_P + F_P} \quad (2)$$

$$\text{Recall} = \frac{T_P}{T_P + F_N} \quad (3)$$

In this context, TP denotes the number of correctly identified dual-stained cervical cells, FP represents the instances where non-dual-stained cells are incorrectly labeled as dual-stained positive, and FN signifies the missed dual-stained positive cells. AP stands for the integral of precision and recall, equating to the area under the Precision-Recall curve, directly reflecting the detection accuracy; a higher AP value indicates better detection performance for a given class. mAP is the average precision across all classes in the dataset. Since this study focuses on detecting a single class, mAP is equivalent to AP in this context.

## 3.3 Results and Discussion

### 3.3.1 Detection Performance

To highlight the advantageous characteristics of our method, we compared the precision (P), recall (R), and mean Average Precision (mAP) of DSIR-YOLO against YOLOv5s, YOLOv5m, YOLOv5l, and YOLOv5x. The suffixes s, m, l, and x denote different versions of the YOLOv5 object detection algorithm, differing primarily in model size and performance capabilities. For instance, YOLOv5n, not mentioned but implied as a comparison point for context, is the smallest version, offering high speed at the expense of detection performance, with a model size of 27.4 MB and computational complexity of 4.5 GFLOPs. Conversely, YOLOv5x, the largest variant, operates at a relatively slower pace but delivers the best detection performance, featuring a model size of 170.5 MB and a computational complexity of 97.8 GFLOPs.

The comparison results, as shown in Table 2, indicate that compared with YOLOv5s, the improved model achieves an increase of 2.0% in precision, 3.8% in recall rate, 7.7% in mAP at IoU threshold 0.5, and 8.9% in mAP at IoU thresholds from 0.5 to 0.95. These results demonstrate that the improvement strategies proposed in this study effectively enhance the recognition accuracy of positive cells in dual-stained images, exhibiting superior performance.

Table 2: Comparison of Detection Performance between This Study and YOLOv5 Series Algorithms

| Model | P/% | R/% | mAP$_{@0.5}$/% | mAP$_{@0.5:0.95}$/% |
|---|---|---|---|---|
| YOLOv5s | 84.8 | 81.1 | 84.9 | 61.6 |
| YOLOv5m | 86.1 | 81.4 | 85.7 | 63.9 |
| YOLOv5l | 86.5 | 83.0 | 86.2 | 64.7 |
| YOLOv5x | 86.5 | 82.3 | 86.4 | 65.5 |
| This Study | 86.8 | 84.9 | 92.6 | 70.5 |

### 3.3.2 Ablation Studies

To validate the performance of each primary module and loss function in the proposed model, ablation experiments were conducted to verify their effectiveness. Using YOLOv5s as the baseline model, nine groups of network models with varying structures were tested to evaluate the impact of different methods on detection performance. The same experimental setup and training parameters were consistently applied across all experiments to assess detection efficacy on the dataset. The outcomes, presented in Table 3, reveal that each improved module in this study contributes to a notable enhancement in the detection accuracy for recognizing positive cells in dual-stained images.

Table 3: Results of Ablation Experiments

| EIoU | Multi-scale | GAM | STR | P/% | R/% | mAP$_{@0.5}$/% | mAP$_{@0.5:0.95}$/% |
|---|---|---|---|---|---|---|---|
| | | Baseline | | 84.8 | 81.1 | 84.9 | 61.6 |
| √ | | | | 83.1 | 81.9 | 88.6 | 63.5 |
| | √ | | | 83.9 | 82.7 | 87.9 | 63.4 |
| | | √ | | 84.9 | 82.4 | 88.1 | 64.0 |
| | | | √ | 84.7 | 82.4 | 88.5 | 64.9 |
| | √ | | √ | 85.7 | 83.5 | 89.5 | 66.5 |
| √ | √ | | √ | 87.1 | 83.0 | 89.7 | 65.5 |
| √ | √ | √ | √ | 85.4 | 84.3 | 91.5 | 69.1 |
| Improvements + Data Augmentation | | | | 86.8 | 84.9 | 92.6 | 70.5 |

The integration of Swin-Transformer effectively captures both local and global information in images, enhancing feature representation capabilities. Specifically, it boosts mAP@0.5:0.95 by 3.3% compared to the baseline network, indicating a significant improvement in detection and recognition accuracy. While this module increases the number of parameters, careful design and optimization can mitigate computational complexity and memory consumption.

EIoU (Enhanced Intersection over Union) provides a better measure of overlap between predicted and ground-truth boxes, evidenced by a 3.7% increase in mAP@0.5 over the baseline, highlighting substantial advancements in localization accuracy. This loss function also accelerates model convergence and mitigates model degradation due to imbalanced positive-negative sample distribution.

Attention mechanisms, embodied by the GAM (Guided Attention Module), improve other metrics without compromising precision. GAM adaptively focuses on useful local information based on spatial relationships in object detection, reducing miss-detection rates, albeit with a slight increase in computational complexity.

Multi-scale feature fusion effectively combines features of varying scales in dual-stain cell images, enhancing perception across different regions. Compared to the baseline, this approach lifts mAP@0.5:0.95 by 1.8%, affirming its contribution to detection accuracy.

The lower half of Table 6-3 details experiments combining multiple strategies, revealing that multi-module utilization outperforms single-module implementations. Building upon Swin-Transformer with multi-scale feature fusion further enhances extraction of effective information in cellular regions, boosting mAP@0.5 by an additional 1%. Incorporating EIoU refines bounding box regression, hastens model convergence, and prevents training degradation due to class imbalance. Finally, adding the GAM module adaptively emphasizes informative local areas to minimize false negatives, albeit with some added computational overhead. This ensemble of strategies, coupled with data augmentation, elevates dual-stained positive cell detection, reinforcing the model's robustness through richer feature information.

### 3.3.3 Comparison with Other Algorithms

To objectively assess our model's detection efficacy, comparative experiments were conducted against existing object detection methodologies, including the classic Faster R-CNN and SSD, the recently published high-performance model YOLOX, and the state-of-the-art YOLOv8. For fair comparison, mean Average Precision (mAP), along with model size (Params) and computational cost measured in Gigaflops per second (GFLOPs), were employed. Table 4 summarizes the results of these comparative experiments.

Table 4: Performance Comparison of Different Algorithms

| Model | mAP$_{@0.5}$/% | mAP$_{@0.5:0.95}$/% | Params/M | GFLOPs |
| --- | --- | --- | --- | --- |
| YOLOv5s | 84.9 | 61.6 | 7.2 | 16.5 |
| YOLOXs | 85.5 | 57.9 | 9.0 | 26.8 |
| YOLOv8s | 87.5 | 64.8 | 11.2 | 28.6 |
| Faster R-CNN | 89.2 | 63.2 | 29.2 | 57.2 |
| SSD512 | 87.5 | 58.4 | 27.2 | 180.6 |
| This Study | 92.6 | 70.5 | 10.6 | 62.9 |

The standard YOLOv5s, while being minimal in terms of parameter count and floating-point operations (FLOPs), also delivers the least effective detection performance. YOLOX and YOLOv8, employing their small (s) models, show improvements in computational demands compared to YOLOv5, evidencing their more advanced architectures. The increase in parameter counts also contributes to enhanced detection accuracy for these models. Conversely, Faster R-CNN, a representative Two-Stage detector, incurs higher parameter and FLOP requirements than the YOLO variants, yet it achieves superior recognition capabilities, particularly with the mAP metric. The SSD512, despite its substantial computational load, fails to demonstrate proportional performance gains.

This study's augmented YOLOv5s incorporates various modules, leading to an increase in parameters compared to the standard version. The integration of the Swin-Transformer module significantly escalates the computational cost, thereby sacrificing some of the original model's lightweight simplicity. This increase in computation also diminishes inference speed, but concurrently, it achieves a substantial boost in mean Average Precision (mAP), enabling better detection of small cells in dual-stained images.

Relative to the baseline models, the magnitude of improvement across different mAP thresholds (mAP@0.5 and mAP@0.5:0.95) varies. For instance, Faster R-CNN demonstrates a greater uplift in mAP@0.5 over YOLOv8s, suggesting superior object localization capabilities. Faster R-CNN is more adept at predicting object locations accurately, where Intersection over Union (IoU) exceeds 0.5. On the other hand, YOLOv8s excels in mAP@0.5:0.95, implying consistent performance across a range of IoU thresholds. This potentially signifies that YOLOv8s offers more precise object localization, maintaining high performance even with stricter demands on bounding box precision. It might also reflect differential handling of objects of varying sizes, as slight deviations in predicted boxes for small objects can drastically reduce IoU values. Hence, DSIR-YOLO is particularly well-suited for dual-stained images, effectively capturing positive cell positions, leveraging the benefits of attention mechanisms.

### 3.3.4 Five-Fold Cross-Validation

Given the scarcity of cervical cell images in the dataset, simple cross-validation may introduce randomness that inadequately substantiates the performance improvement of the refined model over the baseline. To address this, we employed a five-fold cross-validation strategy to robustly assess algorithmic performance. As detailed in Table 5, the entire dataset is first randomly partitioned into five equal subsets. Sequentially, one subset serves as the validation set while the remaining four form the training set. This process iterates five times, with each iteration utilizing a different subset as the validation fold. Ultimately, the mean accuracy of the five trained models represents the final accuracy of the model, thereby mitigating the potential influence of random data splits and providing a comprehensive evaluation of the model's effectiveness.

Table 5: Results of Five-Fold Cross-Validation

| Fold | Model | P/% | R/% | mAP$_{@0.5}$/% | mAP$_{@0.5:0.95}$/% |
|---|---|---|---|---|---|
| Fold1 | YOLOv5s | 84.6 | 80.5 | 88.0 | 61.2 |
|  | This Study | 86.9 | 84.6 | 92.3 | 69.2 |
| Fold2 | YOLOv5s | 85.0 | 80.6 | 88.6 | 61.6 |
|  | This Study | 86.9 | 84.4 | 92.4 | 69.6 |
| Fold3 | YOLOv5s | 85.4 | 80.4 | 88.3 | 62.1 |
|  | This Study | 87.4 | 85.3 | 92.7 | 70.2 |
| Fold4 | YOLOv5s | 84.8 | 78.5 | 84.7 | 61.0 |
|  | This Study | 87.0 | 84.0 | 92.0 | 69.1 |
| Fold5 | YOLOv5s | 84.4 | 79.7 | 87.0 | 60.6 |
|  | This Study | 86.6 | 83.9 | 91.7 | 69.2 |
| Mean | YOLOv5s | 84.6 | 80.5 | 88.0 | 61.2 |
|  | This Study | 86.9 | 84.6 | 92.3 | 69.2 |
| Var | YOLOv5s | 0.17 | 0.90 | 3.15 | 0.44 |
|  | This Study | 0.11 | 0.41 | 0.19 | 0.25 |
| $p$ | --- | $p<0.001$ | $p<0.001$ | $p<0.006$ | $p<0.001$ |

Through five-fold cross-validation, we have observed that DSIR-YOLO significantly outperforms the original YOLOv5s model across all evaluation metrics, including precision (P), recall (R), mAP@0.5, and mAP@0.5:0.95 (all $p<0.006$). The notable variance differences further attest to the stability of model performance. Variance, as a statistical measure of data fluctuation, reflects the consistency of model outputs. A stable output from the model ensures predictability and consistency across varying environments and conditions, which is vital for future clinical deployments. The marked variance discrepancy suggests enhanced performance stability under different conditions, thereby minimizing the impact of random errors. These results not only highlight the superior performance of the refined model but also underscore its algorithmic stability and reliability. Our study adopts the mean of the five-fold cross-validation results as the ultimate performance indicator for the model. We look forward to further optimizing and refining our model in future endeavors to better serve practical applications.

### 3.4 Evaluation of Recognition Effectiveness

Figures 11 through 13 illustrate DSIR-YOLO in cell detection and recognition tasks after enhancements. As depicted in Figure 6-1, cervical dual-stained cell images often contain numerous small positive cells. While the YOLOv5s model can identify most positive cells, it struggles with small target recognition.

In Figure 11, the arrow points to an instance where DSIR-YOLO demonstrates enhanced capability in detecting small-sized positive cells, achieving more efficient recognition. Thanks to the computational advantages of the Swin Transformer, along with the integration of multi-scale feature fusion and attention mechanisms, the model becomes better suited to dual-stained cell images, augmenting the extraction of detailed information from small targets and improving the network's ability to detect objects of various sizes.

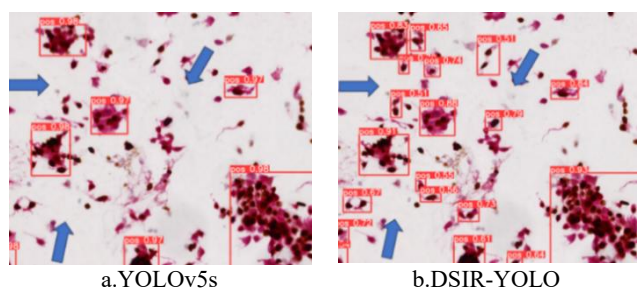

a.YOLOv5s　　　　　　b.DSIR-YOLO

Figure 11: Enhanced Detection of Small Positive Cells

Figure 12 showcases a case where the nucleus and cytoplasm of a cell are not colocated, with the nucleus exhibiting a blue-purple hue, which does not meet the criteria for p16/Ki-67 dual-stain positivity, indicating a false-positive cell. Comparison reveals that the refined model successfully avoids identifying such false-positive cells. The introduction of the attention module enables the model to distinguish between positive and negative cell features more accurately.

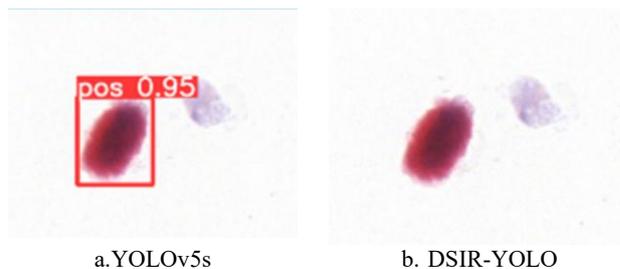

a.YOLOv5s　　　　　　b. DSIR-YOLO

Figure 12: Reduction in False Positive Misidentifications

Figure 13 illustrates advancements in the improved model's reduction of true positive misses and enhancement of bounding box accuracy. In the bottom right, indicated by the arrow, the refined model detects cell regions overlooked by YOLOv5s, indicating more precise positive cell identification. On the left, where the arrow points, DSIR-YOLO accurately delineates clusters of cells, avoiding the overlapping bounding box issue seen in the YOLOv5s model. Multi-scale feature fusion endows the model with stronger feature extraction capabilities, enabling more effective recognition of positive cell clusters in dual-stained cell images.

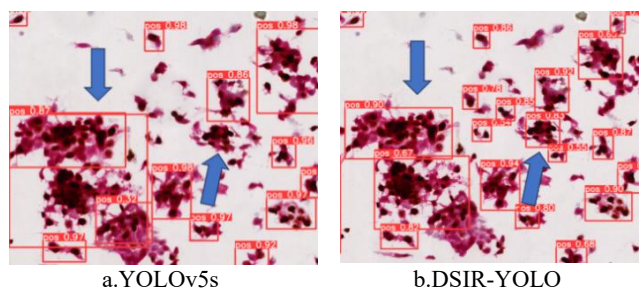

a.YOLOv5s　　　　　　b.DSIR-YOLO

Figure 13: Reduced Missed Detections and Improved Bounding Box Precision

## 4 Conclusion

This paper focuses on liquid-based dual-stained cervical cell images, investigating an identification algorithm based on an improved YOLOv5 for detecting p16/Ki-67 dual-stained positive cells in liquid-based cervical cytology images. Building upon the current state of digital pathology and computer-aided diagnosis research, the algorithm achieves a commendable detection performance for dual-stained positive cervical cell images, thereby assisting in the diagnosis of precancerous and cancerous lesions of the cervix.

In the cellular recognition model, the quality of dataset annotations directly impacts recognition effectiveness. By controlling pixel encapsulation, scale disparities, unannotated cells, and diagonal annotations, annotation noise and inconsistencies were minimized, reducing noise and bias and thereby enhancing the robustness and generalization of the model. Experimental validation and comparisons confirmed that enhancing the annotation quality of the dataset led to a 13.3% increase in precision, a 15.3% boost in recall, an 18.3% improvement in mAP@0.5, and a significant 30.5% uplift in mAP@0.5:0.95.

Focusing on the characteristics of liquid-based dual-stained cervical cell images, DSIR-YOLO by incorporating Swin-Transformer modules, GAM attention mechanisms, multi-scale feature fusion, and the EIoU loss function. These enhancements bolster the model's feature extraction capabilities, addressing the inadequacy of the original network in detecting small-sized cell targets in dual-stained pathological images. Simultaneously, they reinforce the model's capacity to fuse features from both low-resolution and high-resolution scales, mitigating scale discrepancies arising from cells of varying sizes within images, thus enhancing overall accuracy. Results from the five-fold cross-validation experiments indicate post-optimization improvements, raising precision to 86.9%, recall to 84.6%, and mAP@0.5 to 92.3% on the test set of cellular images.